\documentclass[conference,a4paper]{IEEEtran}
\IEEEoverridecommandlockouts

\usepackage[hidelinks]{hyperref}
\usepackage[cmex10]{amsmath}
\usepackage{amssymb,amsfonts}
\usepackage{dblfloatfix}

\usepackage[ruled,vlined]{algorithm2e}
\usepackage{graphicx}
\graphicspath{{Figures/PDF/}{Figures/PNG/}}

\usepackage{booktabs}
\usepackage{siunitx}
\usepackage[numbers,compress]{natbib}
\usepackage{texnames}
\usepackage{bm,bbm}
\usepackage{orcidlink}
\usepackage{lettrine}
 
\begin{document}
\title{PiCSRL: Physics-Informed Contextual Spectral Reinforcement Learning}
\author{
    \IEEEauthorblockN{
        Mitra Nasr Azadani\orcidlink{0009-0001-3508-545X},
        Syed Usama Imtiaz\orcidlink{0000-0002-1229-0100}, and
        Nasrin Alamdari\orcidlink{0000-0001-6595-6669}
    }
    \IEEEauthorblockA{
        \textit{Department of Civil and Environmental Engineering
        Florida State University, Tallahassee, FL, USA}\\
        \{mn22, si22j, nalamdari\}@fsu.edu
    }

   \thanks{Copyright 2026 IEEE. Published in the 2026 IEEE International Geoscience and Remote Sensing Symposium (IGARSS 2026), scheduled for 9 - 14 August 2026 in Washington, D.C.. Personal use of this material is permitted. However, permission to reprint/republish this material for advertising or promotional purposes or for creating new collective works for resale or redistribution to servers or lists, or to reuse any copyrighted component of this work in other works, must be obtained from the IEEE. Contact: Manager, Copyrights and Permissions / IEEE Service Center / 445 Hoes Lane / P.O. Box 1331 / Piscataway, NJ 08855-1331, USA. Telephone: + Intl. 908-562-3966.}
}

\maketitle
\begin{abstract}
High-dimensional low-sample-size (HDLSS) datasets constrain reliable environmental model development, where labeled data remain sparse. Reinforcement learning (RL)-based adaptive sensing methods can learn optimal sampling policies, yet their application is severely limited in HDLSS contexts. In this work, we present PiCSRL (Physics-Informed Contextual Spectral Reinforcement Learning), where embeddings are designed using domain knowledge and parsed directly into the RL state representation for improved adaptive sensing. We developed an uncertainty-aware belief model that encodes physics-informed features to improve prediction. As a representative example, we evaluated our approach for \textit{cyanobacterial} gene concentration adaptive sampling task using NASA PACE hyperspectral imagery over Lake Erie. PiCSRL achieves optimal station selection (RMSE = 0.153, 98.4\% bloom detection rate, outperforming random (0.296) and UCB (0.178) RMSE baselines, respectively. Our ablation experiments demonstrate that physics-informed features improve test generalization (0.52 R², +0.11 over raw bands) in semi-supervised learning. In addition, our scalability test shows that PiCSRL scales effectively to large networks (50 stations, $>$2M combinations) with significant improvements over baselines (p = 0.002). We posit PiCSRL as a sample-efficient adaptive sensing method across Earth observation domains for improved observation-to-target mapping.
\end{abstract}

\title{PiCSRL: Physics-Informed Contextual Spectral Reinforcement Learning}

\section{Introduction}

\lettrine{R}{einforcement learning} (RL) based adaptive sensing methods (ADS)~\cite{pecioski2023overview} learn \textit{when} and \textit{how} to sense with in action space (e.g., sampling rate, sensor activation decision) via optimization that maximizes information gain and minimizes resource use. Yet, this learning gets severely constrained in \textit{high-dimensional low-sample-size} (HDLSS) ~\cite{chadebec2023data} contexts, in particular, environmental monitoring. In the realm of Earth observation (EO), remotely sensed (RS) data have driven a paradigm shift to complement ground-truth data \cite{nuriddinov2026flood}. From near-surface sensors to the newly launched NASA's Plankton, Aerosol, Cloud, and Ocean Ecosystem (PACE) mission, which provides hundreds of contiguous spectral bands with increasingly high-resolution data and rich spatial details. However, this unprecedented capability for environmental monitoring (i.e., water quality ~\cite{imtiaz2025simclr, rabby2026application}, precision in agriculture) requires learning from labeled examples, i.e., ground-truth measurements, which remain sparse and limited. In this regard,  when feature dimension approaches or exceeds sample size, the mathematical foundations of learning change fundamentally, and the covariance estimates become unreliable, which makes models fit noise rather than the true signal. ADS dynamically allocates monitoring resources based on belief state to improve detection efficiency \cite{lermusiaux2007adaptive}. When the action space is small, simple heuristic methods may suffice, and optimal solutions can be derived via exhaustive search. However, as the action space increases, the complexity of finding optimal combinations grows exponentially. Current ADS methods in environmental monitoring have explored Gaussian process bandits \cite{srinivas2010gaussian}, information-directed sampling \cite{russo2018learning}, and spatial utility optimization \cite{krause2008near} as solutions. HDLSS constraints are often addressed through regularization techniques that reduce model complexity using dimensionality reduction and data augmentation  \cite{tuia2016domain}. These challenges, with increasingly high action space and HDLSS for \textit{adapted sensing}, are further exacerbated by the inherent nature of the sequential decision-making task in the domain of water quality, which is inherently challenging. Such as constraints imposed by catchment hydrology ~\cite{nasrazadani2025role}, coupled with complex, nonlinear bacterial ecological behavior \cite{ali2026near, alamdari2026algal}. Moreover, in spectral sensing applications where physical laws govern observations \cite{imtiaz2026spectm}, this creates a fundamental \textit{representation problem}. We propose PiCSRL: \textit{Physics-Informed Contextual Spectral Reinforcement Learning}, where embeddings are designed using domain knowledge and parsed directly into the RL state representation for improved adaptive sensing. In a representation example, we validate our approach in policy sampling for \textit{cyanobacteria} genes concentration prediction using NASA PACE hyperspectral imagery over Lake Erie, USA. Our contributions are as follows:

\begin{enumerate}
    \item We posit that PiCSRL bridges HDLSS constraints with an improved representation mechanism for models utilizing hyperspectral data.
    
    \item We introduce PiCSRL, the first RL framework using hyperspectral sensing under HDLSS constraints for sample-efficient policy learning.
    
    \item We demonstrate direct hyperspectral-to-\textit{cyanobacterial} genes predictive modeling and to our knowledge, this is the first \textit{Algal bloom} sequential decision policy RL from hyperspectral imagery for improved lake sampling. 
    
\end{enumerate}

\section{Methods}

\setlength{\abovecaptionskip}{0.25pt}
\setlength{\belowcaptionskip}{0.25pt}
\begin{figure*}[!t]
\centering
\includegraphics[width=0.88\textwidth]{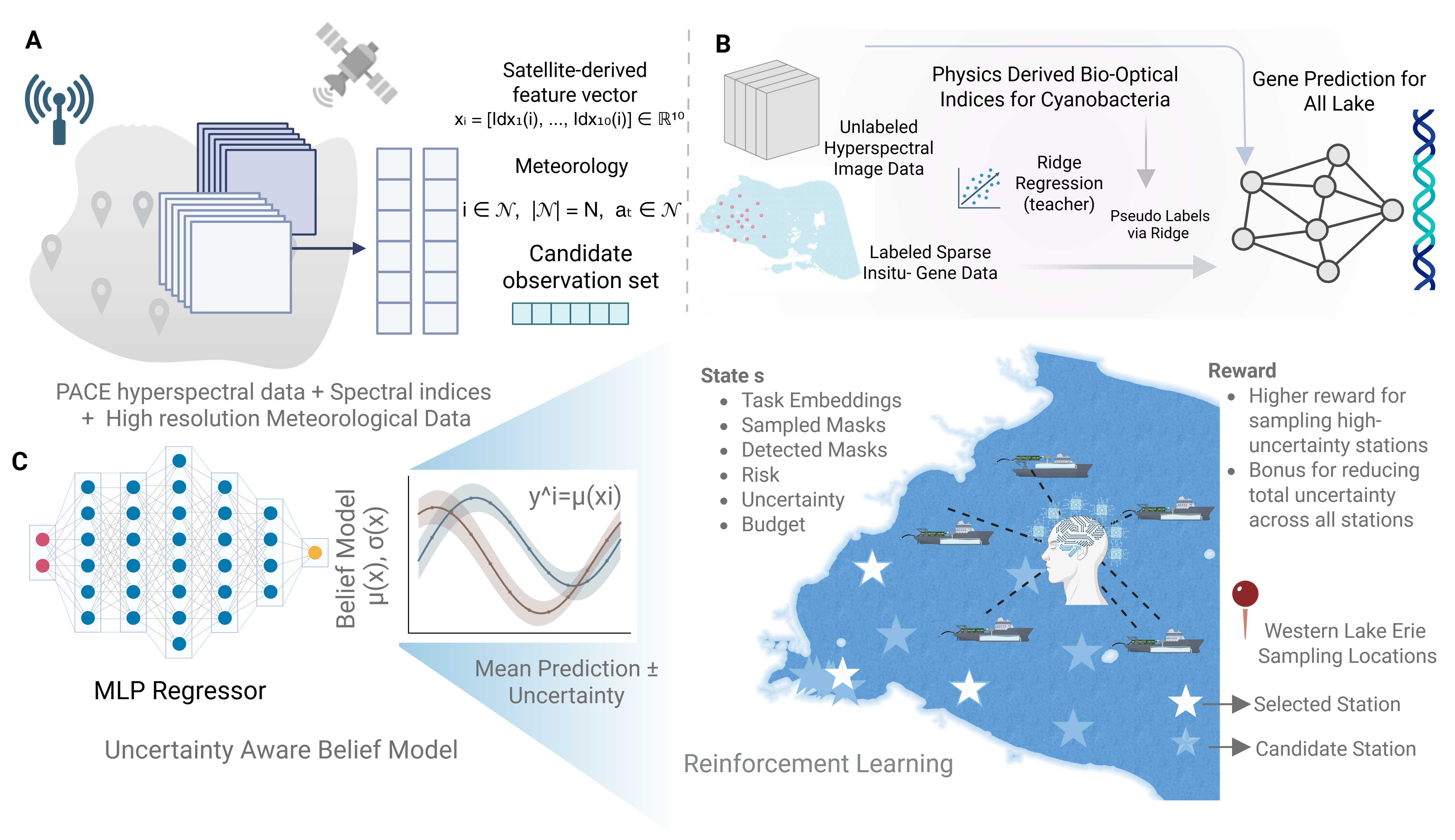}
\vspace{-7pt}
\caption{Physics-Informed Contextual Spectral Reinforcement Learning (PiCSRL) framework. Physics-informed bio-optical indices and sparse in-situ observations are integrated through a semi-supervised learning framework and an uncertainty-aware belief model. These uncertainty-aware predictions are then used in a reduced RL state representation to guide adaptive station selection under sampling constraints.}
\label{fig:PiCSRL_Framework_finalss}
\end{figure*}

\subsection{\textbf{Problem Formulation}}

 Let's $\mathbf{x} \in \mathbb{R}^d$ denote sensor observations with $d$ features and $y \in \mathbb{R}$ the target quantity at each location. The agent has access to $n$ labeled training samples where $d > n$, the HDLSS condition. The task is to learn a policy $\pi$ and sequentially select $K$ locations from $N$ candidates to maximize information about the spatial field subject to budget constraints.


\subsection{\textbf{Physics-Informed Semi-Supervised Learning}}

Let's $\phi: \mathbb{R}^d \rightarrow \mathbb{R}^{d'}$ denote a physics-informed transformation where $d' \ll d$. For our spectral observations, $\phi$ computes ten indices derived from established spectroscopic relationships (Table~\ref{tab:indices}):

\begin{equation}
    \mathbf{x}_i^{\text{phys}} = \phi(\mathbf{r}_i) = [I_1(\mathbf{r}_i), I_2(\mathbf{r}_i), \ldots, I_{10}(\mathbf{r}_i)]^T
\end{equation}

Where $\mathbf{r}_i \in \mathbb{R}^{286}$ is the raw hyperspectral reflectance, and does each index $I_j$ encode a specific physical mechanism. We employ a teacher-student semi-supervised learning (SSL) pipeline to leverage unlabeled observations with dimensionality reduced through physics-informed features. We first train a ridge regression model on the labeled dataset using the physics-informed features:
\begin{equation}
    \hat{y}_i = \mathbf{w}^T \mathbf{x}_i^{\text{phys}} + b, 
    \quad \text{where } \mathbf{x}_i^{\text{phys}} \in \mathbb{R}^{10}
\end{equation}

The regularization parameter $\alpha=1.0$ was selected via 5-fold cross-validation to prevent overfitting on the small labeled set. Ridge regression serves as a robust teacher due to its closed-form solution, resistance to multicollinearity, and well-calibrated predictions under regularization. The trained teacher model generates pseudo-labels for all unlabeled pixels by applying it to their physics-informed features:
\begin{equation}
\tilde{y}_u = f_{\text{teacher}}(\phi(\mathbf{r}_u)),
\quad \forall \mathbf{r}_u \in \mathcal{D}_{\text{unlabeled}}
\end{equation}

Pseudo-labels are constrained to the empirical range $[\min(y_{\text{train}}), \max(y_{\text{train}})]$ to enforce physical plausibility. We adopted a multi-layer perceptron (MLP) serves as the student model, and train it on the combined labeled and pseudo-labeled datasets totaling 60,215 samples. The architecture consists of two hidden layers (64 and 32 neurons) with batch normalization after each layer, ReLU activation, and dropout ($p=0.3$) for regularization. We use a weighted loss function that anchors predictions to ground-truth measurements while leveraging the broader spectral distribution captured in unlabeled data:
\begin{equation}
\mathcal{L} = \frac{1}{N} \sum_{i=1}^{N} w_i (y_i - \hat{y}_i)^2
\end{equation}
Where $w_i = 10$ for labeled samples and $w_i = 1$ for pseudo-labeled samples. This 10:1 weighting ratio was chosen to balance trust in ground-truth versus teacher predictions.

\subsection{\textbf{Reinforcement Learning in Reduced State Space}}
Our belief model employs a shallow neural network for point prediction, together with a bootstrap ensemble for uncertainty estimation. Each network $f_m$ maps transformed features $\mathbf{z} = \phi(\mathbf{x})$ to predictions, with the ensemble providing uncertainty through prediction disagreement:
\begin{equation}
    \mu(\mathbf{z}) = \frac{1}{M}\sum_{m=1}^{M} f_m(\mathbf{z}), \quad \sigma(\mathbf{z}) = \sqrt{\frac{1}{M}\sum_{m=1}^{M}(f_m(\mathbf{z}) - \mu(\mathbf{z}))^2}
\end{equation}
where $M$ denotes ensemble size. Network architecture is kept shallow to prevent overfitting in HDLSS conditions. The state at decision step $t$ comprises predicted values $\boldsymbol{\mu} \in \mathbb{R}^N$ and uncertainties $\boldsymbol{\sigma} \in \mathbb{R}^N$ for candidate locations, along with a binary mask indicating previously visited sites. State dimensionality scales with candidate locations $N$, \textit{independent of raw observation dimension} $d$. Actions correspond to selecting an unvisited candidate location. The reward function balances information acquisition with uncertainty reduction:
\begin{equation}
    r_t = \alpha \cdot r_{\text{info}}(a_t) + \beta \cdot r_{\text{uncert}}(a_t) + \gamma \cdot r_{\text{spatial}}(a_t)
\end{equation}
where rewards for info is the negative absolute prediction error, uncert is the epistemic uncertainty at selected location, and spatial is min distance to previously selected stations; component weights are determined through sensitivity analysis. Policy learning is performed using a deep Q-learning framework trained with uniform experience sampling \cite{wang2016dueling}. Training proceeds through simulated episodes constructed from the belief model predictions, requiring no additional real-world samples beyond those used for belief model construction. This simulation-based training helps model capture the physics-informed state representation.

\begin{table}[t]
\caption{Physics-Based Spectral Indices}
\label{tab:indices}
\centering
\footnotesize
\begin{tabular}{@{}p{0.5cm}p{3.5cm}p{3.2cm}@{}}
\hline
\textbf{Index} & \textbf{Formulation} & \textbf{Physical Mechanism} \\
\hline
CI & $\rho_{681} - \rho_{665}$ & Chlorophyll fluorescence \\
NDCI & $(\rho_{709} - \rho_{665})/(\rho_{709} + \rho_{665})$ & Red-edge response \\
MCI & $\rho_{709} - (\rho_{681} + \rho_{753})/2$ & Maximum chlorophyll \\
FAI & \textit{See \cite{hu2009fai}} & Floating algae index \\
PC & $\rho_{620}/\rho_{665}$ & Phycocyanin absorption \\
ChlRed & $\rho_{680}/\rho_{665}$ & Chlorophyll/red ratio \\
BG & $\rho_{443}/\rho_{555}$ & Blue/green ratio \\
GR & $\rho_{555}/\rho_{665}$ & Green/red ratio \\
NIR & $\rho_{865}/\rho_{665}$ & NIR/red ratio \\
NDI & $(\rho_{665} - \rho_{620})/(\rho_{665} + \rho_{620})$ & Normalized difference \\
\hline
\multicolumn{3}{@{}p{6.8cm}@{}}{\footnotesize Representative indices shown; full set includes ten features. $\rho_\lambda$ denotes reflectance at wavelength $\lambda$ nm. Formulations derive from established bio-optical relationships [4, 5].} \\
\end{tabular}
\end{table}

\section{Experiments}

\subsection{\textbf{Experimental Setup}}

We employ hyperspectral imagery from NASA's PACE Ocean Color Instrument (OCI) eight-day composite products paired with ground-truth measurements (cyanobacteria gene abundance) from Western Lake Erie. Our training data comprises 98 station-days from 2024; testing uses 92 station-days from 2025. These eight monitoring stations span the western basin, and adaptive selection is performed under a fixed sampling budget of three stations per sampling event. For rigorous analysis, we implemented detailed baseline experiments that include random selection, spatially-stratified sampling, greedy intensity-based selection, and an upper confidence bound (UCB) strategy based on belief mean and uncertainty \cite{srinivas2010gaussian}.

\subsection{\textbf{Representation Learning for HDLSS}}

We first analyzed the effect of physics-informed features on generalization using a semi-supervised learning approach. Our results demonstrate that raw spectral features achieve higher training fit but substantially lower test generalization (Test R² = 0.41) compared to physics-informed indices (Test R² = 0.52). In principle, When in-sample performance is higher in comparison to out-of-sample performance, indicates that  model learns spurious correlations specific to training data rather than generalizable relationships. 
Our semi-supervised approach yields modest but consistent improvement over the supervised baseline. The ensemble model achieves a test $R^2 = 0.517$ compared to the teacher's $R^2 = 0.516$, with the 613$\times$ expansion in training data (from 98 to 60,215 samples) providing marginal gains in generalization. The limited improvement suggests that hand-crafted physics-based indices already encode the bloom-relevant spectral signal effectively, with minimal additional information accessible through pseudo-label augmentation. This validates our hypothesis that physics-informed dimensionality reduction is the primary mechanism addressing HDLSS challenges, and SSL provides incremental benefit by expanding coverage of the spectral feature space. 

\begin{table}[t]
\centering
\caption{Generalization Performance by Feature Representation}
\label{tab:hdlss}
\begin{tabular}{@{}lccc@{}}
\toprule
\textbf{Features} & \textbf{Dimension} & \textbf{Train R$^2$} & \textbf{Test R$^2$} \\
\midrule
Physics-Informed Indices & 10 & 0.47 & 0.52 \\
Raw Spectral Bands & 117 & 0.52 & 0.41 \\
Combined & 127 & 0.54 & 0.49 \\
\bottomrule
\end{tabular}
\vspace{-2mm}
\end{table}

\subsection{\textbf{Adaptive Strategy Performance}}

\begin{figure}[t]
    \centering
    \includegraphics[width=\linewidth]{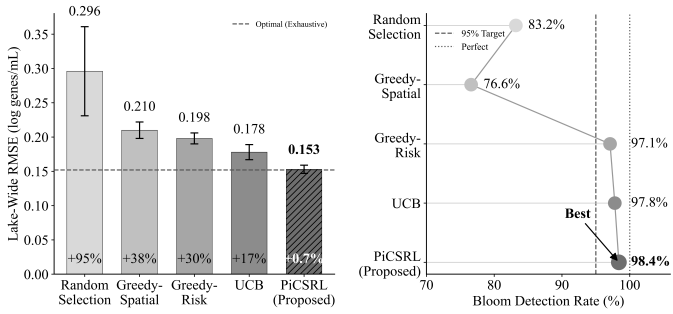}
    \caption{Adaptive sampling performance for selecting $K=3$ stations from $N=8$ candidates. 
    Left: Lake-wide reconstruction error (RMSE; mean $\pm 1\sigma$ over 500 episodes), with the optimal exhaustive baseline shown as a dashed line. Right: Bloom detection rate (\%), with the 95\% operational target and the perfect (100\%) reference indicated.}
    \label{fig:adaptive_sampling}
\end{figure}

We performed adaptive sensing experiment with eight candidate stations and three selection budget with exhaustive enumeration to identify the optimal station combination and direct verification of learned policies. PiCSRL Fig.~\ref{fig:adaptive_sampling} consistently selects the station subset with minimum reconstruction error and commensurate the exhaustive optimum (RMSE = 0.1527 ± 0.006) and effectively attaining optimal performance. In contrast, heuristic baselines exhibit substantially higher reconstruction error, including Greedy-Spatial (RMSE = 0.2098) and Greedy-Risk (RMSE = 0.1982), while random selection performs least (RMSE = 0.2958). Uncertainty-aware selection improves performance relative to purely heuristic algorithms in reducing reconstruction error (RMSE = 0.178 ± 0.011) and increasing bloom detection accuracy. However, the proposed PiCSRL framework achieves the best overall performance in simultaneously minimizing reconstruction error and maximizing detection rate (98.4\%). In addition, the computational comparison further favors the proposed approach, where PiCSRL inference requires only a forward pass through a trained deep Q-network, while exhaustive search requires iterative finding of optimal combinations for station counts and this becomes infeasible for large-scale networks. This efficiency advantages the deployment of RL-based adaptive sensing for large spatial scales to preserve near-optimal performance.

\subsection{\textbf{Scalability Analysis}}

\begin{figure}[t]
    \centering
    \includegraphics[width=\linewidth]{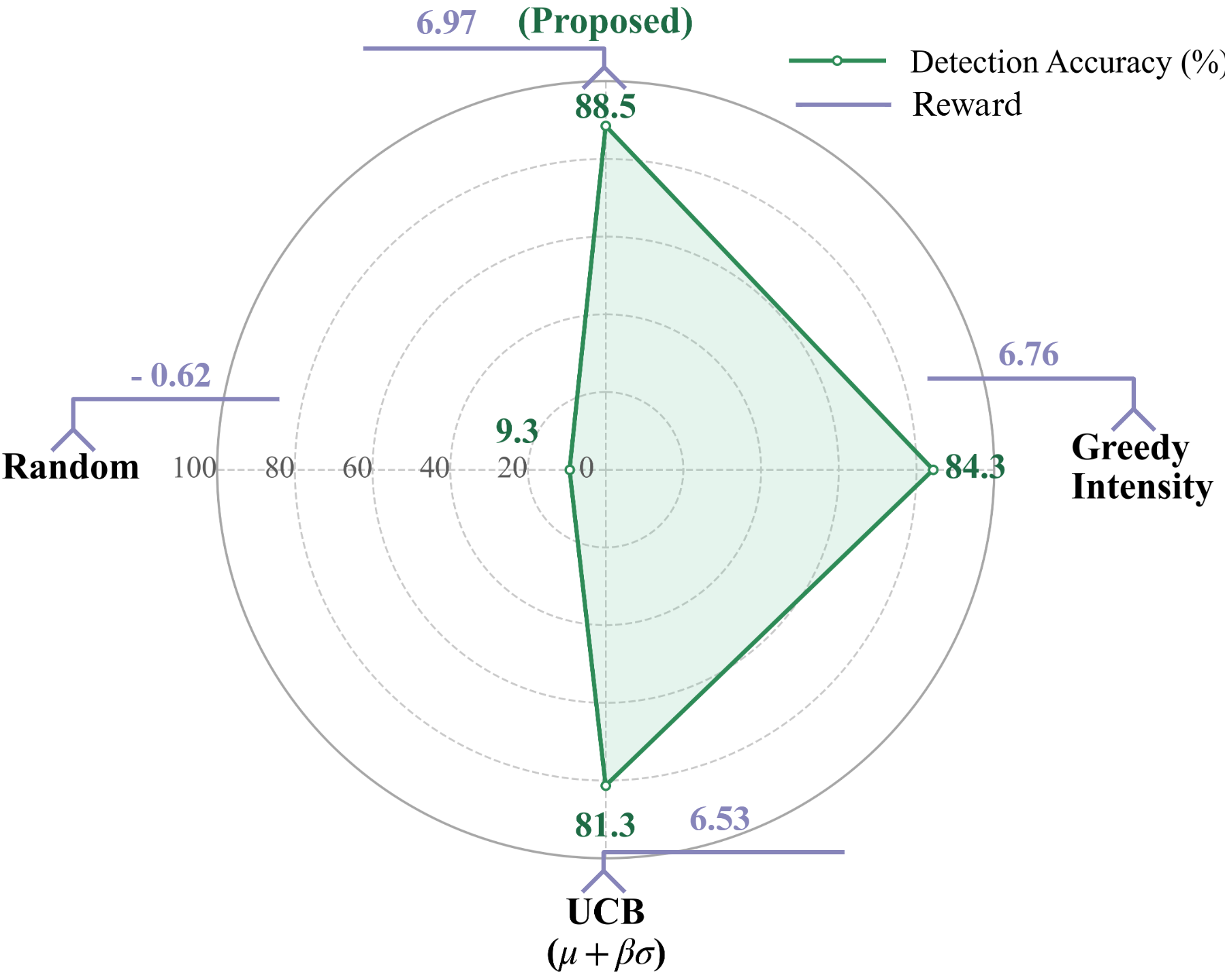}
\caption{Performance comparison of adaptive sampling strategies. 
Left: Detection accuracy for comparison methods and the proposed PiCSRL method. Right: Corresponding cumulative reward achieved by each strategy. PiCSRL attains the highest detection accuracy and cumulative reward, with statistically significant improvement over baseline methods ($p=0.002$).}

    \label{fig:increased_font_figure1}
\end{figure}

In order to evaluate our model performance for large scale deployment, we constructed a 50-station adaptive sensing scenario, where  virtual stations were spatially distributed and formed over two million possible selection combinations. PiCSRL Fig.~\ref{fig:increased_font_figure1} maintains superior performance advantages at this large scale and achieves the highest bloom detection rate (88.5\%) where largest cumulative reward was 6.97. In comparison, Greedy-Risk achieves 84.3\% detection, while uncertainty-based UCB selection reaches 81.3\%. Random selection performs at least with only (9.3\%) detatection rate and reflects the difficulty of the task at scale. Our statistical test confirms that the improvement achieved by PiCSRL over Greedy-Risk is significant (p = 0.002), while the UCB baseline also differs significantly from the reference (p = 0.042). These results further substantiate that the proposed approach scales effectively to large candidate sets.

\section{Discussion}
Our physics-informed RL provides a mechanistic representations that overcome HDLSS data constraints without any explicit reliance on regularization. We note, in ablation experiments, higher-dimensional features achieve better training performance but didn't perform well on test datasets. This is evidence that the model overfitted and learned spurious correlations. In addition, hyperspectral sensors such as PACE have contiguous narrow wavelengths; while this enhances environmental modeling to obtain precise spectral signatures, it also induces multicollinearity when target features span multiple bands. The pretrained embeddings encode physics-based representations for the downstream task; the RL agent does not need to rediscover them from raw data. This substantially reduces complexity, which leads to a faster learning and better decisions. In our experiments, we compare against a UCB baseline $(\mu + \beta \sigma)$ that combines ensemble predictions with uncertainty bonuses. DQN consistently outperforms this heuristic with statistical significance. Unlike Gaussian Process methods that would require cubic computational scaling and suffer from kernel degradation in high dimensions, our bootstrap ensemble provides computationally efficient uncertainty estimates while operating in the physics-informed feature space. This architectural choice, learning policies in a reduced belief space rather than raw observation space, enables RL to succeed in HDLSS conditions where traditional kernel-based methods face fundamental limitations.

Limitations and Future Work: Our current approach requires domain experts to specify physics-informed features, and its limited to aquatic monitoring. Future work involves multi-objective extensions using more datasets and comparing RL performance.

\section{Conclusion}
In this work, we presented a framework based on an uncertainty-aware, multi-objective RL model. As a representative example, we formulated monitoring as a sequential decision problem for adaptive environmental monitoring, which establishes physics-informed representations in high-dimensional, low-sample-size (HDLSS) contexts. For \textit{cyanobacterial} gene prediction, our model uses domain-informed hyperspectral bands that enable sample-efficient learning over large regions. Our experimental analysis demonstrates that RL consistently identifies near-optimal sampling locations and outperforms various competing methods. As environmental systems are highly interconnected that requires sequential decision-making for long-term benefits; our work lays the foundation for leveraging RS-based data to move beyond sparse and static observation strategies.

\section{Acknowledgment}
Fig.~\ref{fig:PiCSRL_Framework_finalss} is created with Biorender \cite{biorender}

\bibliographystyle{IEEEtran}

\begin{thebibliography}{10}

\bibitem{pecioski2023overview}
D. Pecioski, V. Gavriloski, S. Domazetovska, and A. Ignjatovska, ``An overview of reinforcement learning techniques,'' in \textit{Proc. 12th Mediterranean Conf. Embedded Computing (MECO)}, Budva, Montenegro, 2023, pp. 1--4.

\bibitem{chadebec2023data}
C. Chadebec, E. Thibeau-Sutre, N. Burgos, and S. Allassonnière, ``Data augmentation in high dimensional low sample size setting using a geometry-based variational autoencoder,'' \textit{IEEE Trans. Pattern Anal. Mach. Intell.}, vol. 45, no. 3, pp. 2879--2896, Mar. 2023.

\bibitem{nuriddinov2026flood}
A. Nuriddinov, E. Ahmadisharaf, and M. R. Alizadeh, ``High Resolution Flood Extent Detection Using Deep Learning with Random Forest Derived Training Labels,'' \textit{arXiv preprint arXiv:2603.22518}, 2026. Available: https://arxiv.org/abs/2603.22518

\bibitem{imtiaz2025simclr}
S. U. Imtiaz, M. Nasr Azadani, and N. Alamdari, ``SimCLR-enabled wide and deep learning for cyanobacterial bloom prediction from NASA's PACE hyperspectral mission,'' \textit{IEEE Geosci. Remote Sens. Lett.}, vol. 22, pp. 1--5, 2025, Art. no. 1504905.

\bibitem{rabby2026application}
S. H. Rabby, X. Sun, A. M. I. Hafiz, Z. Yan, S. U. Imtiaz, M. Nasr Azadani, M. Pakdehi, A. S. Moumouni, E. Ahmadisharaf, and N. Alamdari, ``Application of machine learning methods in water quality modeling,'' in \textit{Machine Learning and Artificial Intelligence in Toxicology and Environmental Health}, Z. Lin and W.-C. Chou, Eds. Academic Press, 2026, pp. 271--309.

\bibitem{lermusiaux2007adaptive}
P.~F. Lermusiaux, ``Adaptive modeling, adaptive data assimilation and adaptive sampling,'' \textit{Physica D}, vol.~230, pp.~172--196, 2007.

\bibitem{srinivas2010gaussian}
N.~Srinivas \textit{et al.}, ``Gaussian process optimization in the bandit setting: No regret and experimental design,'' in \textit{Proc. ICML}, pp.~1015--1022, 2010.
\bibitem{russo2018learning}
D.~Russo and B.~Van Roy, ``Learning to optimize via information-directed sampling,'' \textit{Oper.\ Res.}, vol.~66, no.~1, pp.~230--252, 2018.

\bibitem{krause2008near}
A.~Krause \textit{et al.}, ``Near-optimal sensor placements in Gaussian processes: Theory, efficient algorithms and empirical studies,'' \textit{J. Mach. Learn. Res.}, vol.~9, pp.~235--284, 2008.

\bibitem{tuia2016domain}
D.~Tuia \textit{et al.}, ``Domain adaptation for the classification of remote sensing data,'' \textit{IEEE Geosci. Remote Sens. Mag.}, vol.~4, no.~2, pp.~41--57, 2016.

\bibitem{nasrazadani2025role}
M. Nasr Azadani, S. U. Imtiaz, and N. Alamdari, ``Role of impoundment and irrigation in intensive agriculture watersheds,'' \textit{J. Hydrol.}, vol. 662, pt. C, 2025, Art. no. 134075.

\bibitem{ali2026near}
M. A. Salou, S. U. Imtiaz, M. Nasr Azadani, and N. Alamdari, ``Near real-time and next-day prediction for \textit{Escherichia coli} (\textit{E. coli}) concentrations in highly urbanized watersheds,'' \textit{Water Res.}, vol. 290, 2026, Art. no. 125030.


\bibitem{alamdari2026algal}
N. Alamdari, Z. Yan, M. Nasr Azadani, and S. U. Imtiaz, ``Algal blooms,'' in \textit{Data-Driven Earth Observation for Disaster Management}, X. Huang, S. Wang, K. Kalogeropoulos, and A. Tsatsaris, Eds. Elsevier, 2026, pp. 183--205.


\bibitem{imtiaz2026spectm}
S. U. Imtiaz, M. Nasr Azadani, and N. Alamdari, ``SpecTM: Spectral Targeted Masking for Trustworthy Foundation Models,'' \textit{arXiv preprint arXiv:2603.22097}, 2026. Available: https://arxiv.org/abs/2603.22097

\bibitem{hu2009fai}
C.~Hu,
``A novel ocean color index to detect floating algae in the global oceans,''
\textit{Remote Sens. Environ.}, vol.~113, no.~10, pp.~2118--2129, 2009.

\bibitem{wang2016dueling}
Z.~Wang \textit{et al.}, ``Dueling network architectures for deep reinforcement learning,'' in \textit{Proc. ICML}, pp.~1995--2003, 2016.

\bibitem{biorender}
Created in BioRender. Imtiaz, S. U. (2026) https://BioRender.com/q746gxk


\end{thebibliography}

\end{document}